\documentclass{article}

\usepackage{arxiv}

\usepackage[utf8]{inputenc} 
\usepackage[T1]{fontenc}    
\usepackage{hyperref}       
\usepackage{url}            
\usepackage{booktabs}       
\usepackage{amsfonts}       
\usepackage{nicefrac}       
\usepackage{microtype}      

\usepackage{graphicx}       
\usepackage{wrapfig}
\usepackage{subcaption}

\title{World Knowledge from AI Image Generation for Robot Control}

\author{
 Jonas Krumme \\
  Cognitive Neuroinformatics \\
  University of Bremen \\
  Bremen, Germany \\
  \texttt{jkrumme@uni-bremen.de} \\
  \And
  Christoph Zetzsche \\
  Cognitive Neuroinformatics \\
  University of Bremen \\
  Bremen, Germany \\
}

\renewcommand{\headeright}{ }
\renewcommand{\undertitle}{ }

\begin{document}
  \date{}
  \maketitle

\begin{abstract}
  When interacting with the world robots face a number of difficult questions, having to make decisions when given under-specified tasks where they need to make choices, often without clearly defined right and wrong answers. Humans, on the other hand, can often rely on their knowledge and experience to fill in the gaps. For example, the simple task of organizing newly bought produce into the fridge involves deciding where to put each thing individually, how to arrange them together meaningfully, e.g. putting related things together, all while there is no clear right and wrong way to accomplish this task. We could encode all this information on how to do such things explicitly into the robots' knowledge base, but this can quickly become overwhelming, considering the number of potential tasks and circumstances the robot could encounter. However, images of the real world often implicitly encode answers to such questions and can show which configurations of objects are meaningful or are usually used by humans. An image of a full fridge can give a lot of information about how things are usually arranged in relation to each other and the full fridge at large. Modern generative systems are capable of generating plausible images of the real world and can be conditioned on the environment in which the robot operates. Here we investigate the idea of using the implicit knowledge about the world of modern generative AI systems given by their ability to generate convincing images of the real world to solve under-specified tasks.
\end{abstract}


\section{Introduction}

\begin{wrapfigure}{l}{0.50\columnwidth}
  \begin{subfigure}{.48\linewidth}
    \includegraphics[width=\linewidth]{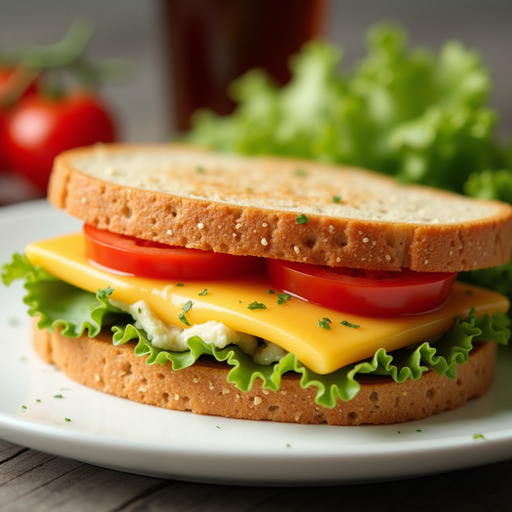}
    \caption{Conventional Stacking}
    \label{fig:sandwich_correct}
  \end{subfigure}
  \begin{subfigure}{.48\linewidth}
    \includegraphics[width=\linewidth]{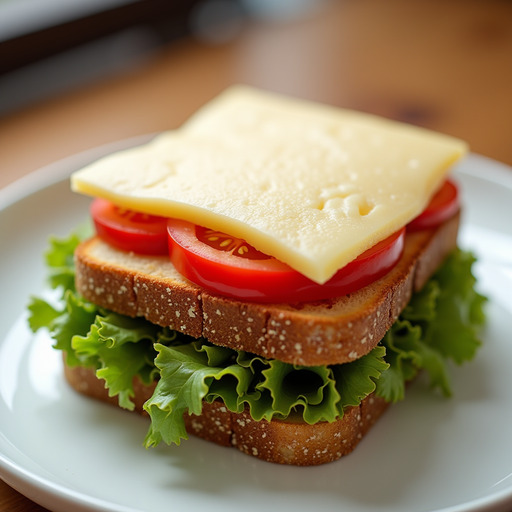}
    \caption{Unconventional Stacking}
    \label{fig:sandwich_wrong}
  \end{subfigure}
  \caption{Two different versions of how to stack the ingredients of a sandwich, where version \textbf{(a)} would likely be seen as the correct version while \textbf{(b)} would be seen as at least unconventional (generated with FLUX.1[dev])}
  \label{fig:sandwich_example}
\end{wrapfigure}

Real images encode a lot of information about the world, such as how an object can look like, how certain things can be meaningfully arranged, or which items belong together. The image of an average office desk can give us information about how the different parts are usually arranged in relation to each other, e.g. a monitor on the desk with mouse and keyboard in front of it and a chair in front of the desk, or the image of someone preparing a meal can give us information about which ingredients and kitchen tools are to be used. This might seem rather trivial from a human perspective as we are very easily capable of handling such tasks without having to rely on pre-made example images to follow, but for a robot that has to navigate and solve tasks in e.g. a household environment such information can be critical for successfully handling everyday-activities and interacting with the world. We could encode all relevant information explicitly into an extensive knowledge base~\cite{lenat1995cyc} for the robot, but considering the number of tasks and circumstances that a robot could encounter, correctly handling all situations could become very challenging~\cite{paulius2019survey} or even overwhelming when the robot needs to act in widely different environments. Additional knowledge sources, such as simulations of the environment, when available, can help by providing ways to investigate consequences of actions without having to act in the world~\cite{beetz2018know}. We could also try to train the robot on a variety of different tasks, e.g. using reinforcement learning or other methods~\cite{li2021igibson}, hoping that the robot is able to generalize and handle situations and circumstances that were never seen during training. However, images of the real world already show examples of how a dining table looks like with plates and cutlery, how images are hung on the wall in bedrooms, dining rooms, or other places. Figure \ref{fig:sandwich_example} shows an example of two different versions of how sandwich ingredients could be stacked together. When given the task of preparing a sandwich using bread, cheese, salad, and tomatoes, a robot without explicit knowledge of how to stack these ingredients might see both variants as valid solutions to the tasks. However, having access to images of realistic sandwiches can reveal to the robot that under normal circumstances \ref{fig:sandwich_correct} would be the expected solution. 
However, just taking images from the Internet or other sources still poses a big challenge when it comes to transferring the knowledge to the exact circumstances of the robot, as it would be unlikely that the layout of the scene in those images matches the layout of the current environment. Access to example images with the same layout as the environment would allow us to directly apply the knowledge and information stored in the images to help the robot solve different tasks~\cite{kapelyukh2023dall}.

\begin{figure}[ht]
	\centering
	\includegraphics[width=\linewidth]{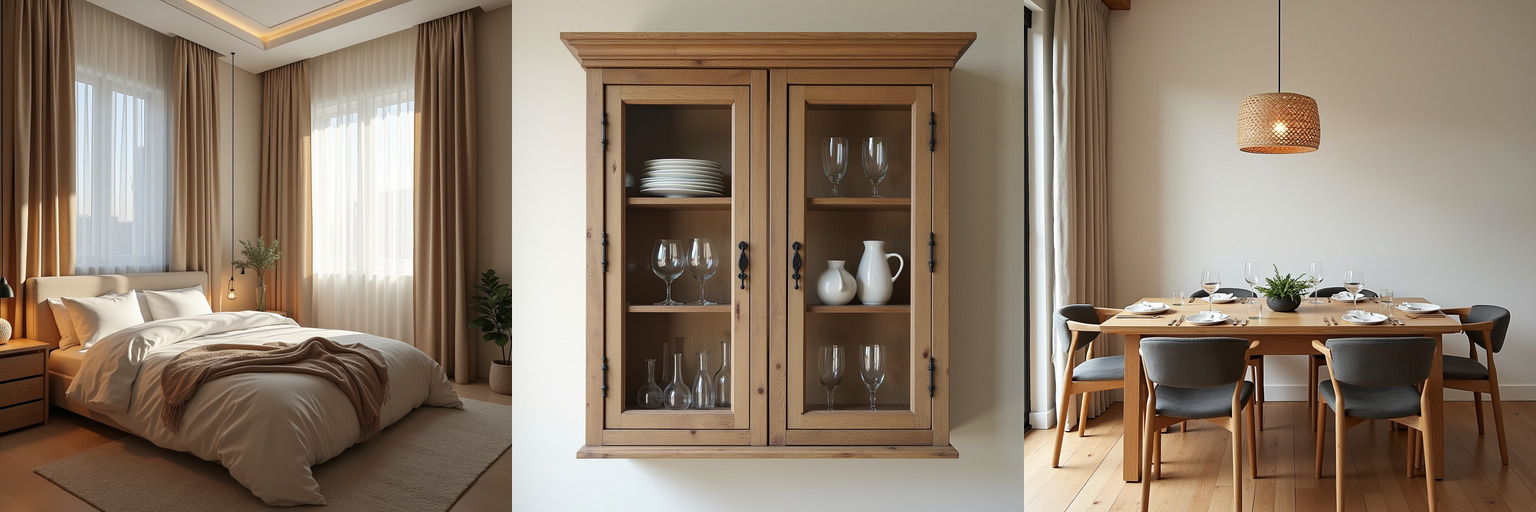}
	\caption{Generated images of different household scenarios that a robot could encounter while navigating a household and potentially performing different tasks (generated with FLUX.1[dev])} 
    \label{fig:generated_household_scenes}
\end{figure}

Image generation models like StableDiffusion~\cite{rombach2022stable1,esser2024stable3}, Dall-E~\cite{ramesh2021zero} or FLUX.1~\cite{Flux} see a large number of images of the real world during training and recent systems are capable of creating convincing images that can fool the naive viewer with only minor artifacts indicating that such images are actually showing artificially generated scenes. Figure \ref{fig:generated_household_scenes} shows some example scenes generated using FLUX.1[dev], a state-of-the-art text-to-image generation model. The images display different scenarios, here inside a house, which a robot could encounter while navigating and performing different tasks in a household environment. We investigated the idea of using open source and/or freely available generative AI systems like the StableDiffusion-family of text-to-image models or the Flux.1[dev] version of the FLUX.1-family, for which the network weights are available, and their capabilities of generating images for solving under-specified robot tasks. 

\section{Related Work}

\subsection{Generative Artificial Intelligence for Image Generation}

Recent progress in the area of image generation with models and model families like Stable Diffusion~\cite{rombach2022stable1,podell2023sdxl,esser2024stable3}, Dall-E~\cite{ramesh2021zero,betker2023improving}, Imagen~\cite{saharia2022photorealistic} or most recently FLUX.1~\cite{Flux} expanded the state of the art in text-to-image generation and image generation in general. Most of these model families went through multiple iterations, improving output quality, prompt adherence, and overcoming shortcomings of previous models. As diffusion~\cite{sohl2015deep} or diffusion-based models have become somewhat of a standard choice for generating high-resolution images, as can be seen with the previously mentioned examples, generative models like GigaGAN~\cite{kang2023gigagan} show that other approaches can also be scaled up for large-scale image generation tasks. Extensions like ControlNet~\cite{zhang2023adding} were developed, giving the user more control over generated images by using depth or edge maps, as well as other modalities. Availability of the model weights for some models, e.g. notably the StableDiffusion-family, in combination with methods like Low-Rank Adaption (LoRA)~\cite{hu2022lora} also allowed larger groups of users to finetune and adapt large image generation models to their own use cases.

Large language models (LLMs)~\cite{zhao2023survey}, among the most notable successes of generative AI in the recent past, have added the capacity to process multiple modalities in addition to text. Recent models are capable of analyzing image and video data~\cite{zhang2024mm} and can be used in combination with dedicated image generation models to generate images based on user requests to the LLM~\cite{koh2023generating}.

\subsection{Generative Models for Robot Control}

Generative models are already used for a wide variety of tasks in the area of robotics. Image generation models are used to generate and enhance training data by generating variations of existing training images, improving the generalization capabilities of robotic systems~\cite{chen2023genaug,yu2023scaling}. Shridhar et al.~\cite{shridhar2024generative} use a finetuned version of Stable Diffusion to generate target images for robot joints, using the image generation system to directly infer the position of the robot joints for a given task. Kapelyukh et al.~\cite{kapelyukh2023dall} use Dall-E to generate target or goal images for object rearrangement tasks. In comparison to our work, they restrict their system to a top-down view of a table where objects are arranged on a flat 2D-plane. In~\cite{kapelyukh2024dream2real} they extend this to more complex 3D object rearrangement tasks by generating goal states with the help of 3D neural radiance fields.

With the advent of more powerful text-to-video or image-to-video models, using information from videos to inform robot actions becomes a more usable prospect for the future. Bharadhwaj et al.~\cite{bharadhwaj2024gen2act} use video generation models to generate videos showing robot motions to a specially trained neural network that tries to follow these actions to accomplish tasks in real-world scenarios.

Another area for enhancing robot actions using generative AI involves the growing development of multimodal LLMs or vision language models (VLMs) that combine the reasoning capabilities of their text-based predecessors with the ability to understand images or other modalities. Driess et al.~\cite{driess2023palme} show a large language model that incorporates sensor data from the robot into its decision making process and can give commands that are executed by the robot control mechanisms. 

\section{Method}

\begin{figure}[ht]
	\centering
	\includegraphics[width=\linewidth]{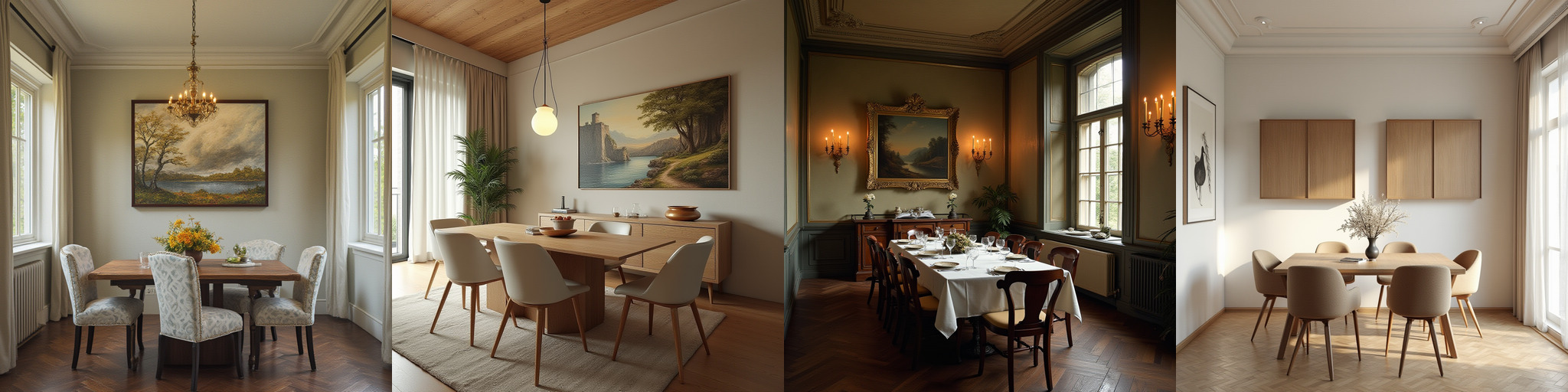}
	\caption{Example images of dining rooms with different layouts and painting positions (generated with FLUX.1[dev])} 
    \label{fig:dining_table_examples}
\end{figure}

A problem with using already existing images found online or taken from other sources is that it is unlikely that they can be directly used to inform robot action. The image might show a dining room with a painting on the wall, but the room might have different dimensions or a different layout in comparison to the room where the robot currently tries to hang up a painting. Furniture may block areas in the image that are available in the actual room. Such divergences could make it impossible to simply use the positioning from the image as the target state for the robot without further preprocessing, excessive search through a large number of images, or more complex work to try to generalize the information found in the given images to the current situation. Figure \ref{fig:dining_table_examples} shows some generated examples of dining rooms with paintings on the wall that have different layouts and dimensions. A robot might find that none of these images is useful in solving the problem in the current environment.

However, if we had a technique to find or generate images of the requested scene with the same overall layout, we would be able to directly use the information from these images to act. We could directly derive the target position from the image and hang the painting on the wall accordingly. This is where modern generative AI systems can help by being able to generate images of the goal state when no existing images matching the current scenario can be found or such a search would be too costly to attempt.

\subsection{Overview}

\begin{figure}[ht]
	\centering
	\begin{subfigure}{.32\linewidth}
		\includegraphics[width=\linewidth]{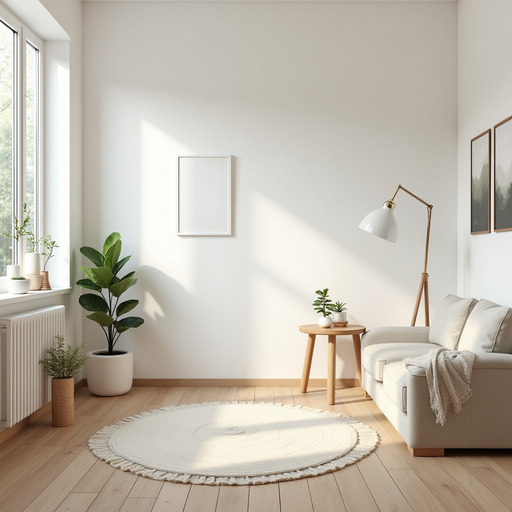}
        \caption{Base Image}
        \label{fig:original_image}
	\end{subfigure}  
	\begin{subfigure}{.32\linewidth}
		\includegraphics[width=\linewidth]{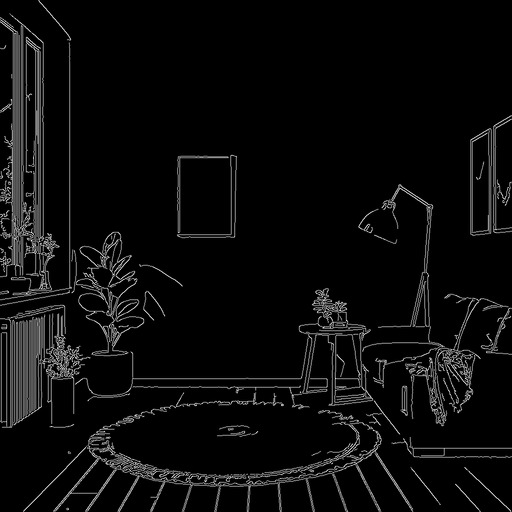}
        \caption{Edge Map}
        \label{fig:canny_image}
	\end{subfigure}
	\begin{subfigure}{.32\linewidth}
		\includegraphics[width=\linewidth]{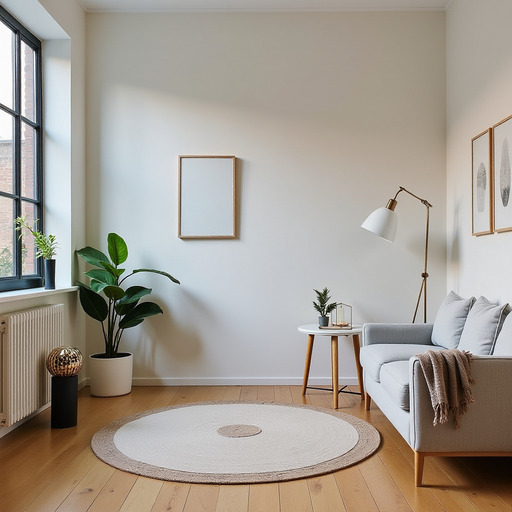}
        \caption{Recreated Image}
        \label{fig:recreated_canny}
	\end{subfigure}
	\caption{Generating images with the same layout from a base image using edge maps. \textbf{(a)} Base image for generating the edge map (generated with FLUX.1[dev]), \textbf{(b)} Edge map from the base image, \textbf{(c)} Image generated using the layout from the edge map (generated with FLUX.1 Canny[dev])} 
    \label{fig:control_net_example}
\end{figure}

To make use of the implicit knowledge given by the ability of modern generative AI systems to generate or reproduce realistic images for guiding robots, we need the ability to condition the system on the current layout of the environment in which the robot is currently acting in. A simple way for text-to-image models would be a textual description of the environment. But ambiguities in such descriptions can easily lead to deviations from the actual layout as seen from the robot view. Kapelyukh et al.~\cite{kapelyukh2023dall} use inpainting and mask out existing objects that should be preserved during image generation. Only a smaller part of the final image is filled in by the generative system. However, this would imply that we need to know the valid regions relevant to the task for the image masks for a variety of circumstances. Here we decided to use ideas introduced by ControlNet~\cite{zhang2023adding} to guide the layout of the generated scene using edge maps calculated from the original image. This allows us to preserve the layout of the scene while the generative system is not restricted to parts of the image when it comes to inserting or arranging new objects. This also removes the need for further preprocessing to find potential image sections for masking. The text prompt is used for further guidance and to add the missing objects that are not given by the edge map or to otherwise enrich the current view of the robot suitable to accomplishing the given task. Figure \ref{fig:control_net_example} shows an example in which a base image is used to create an edge map that guides the image generation process. FLUX.1 Canny[dev] was used to create \ref{fig:recreated_canny}.

\begin{figure}[ht]
  \centering
  \includegraphics[width=.8\linewidth]{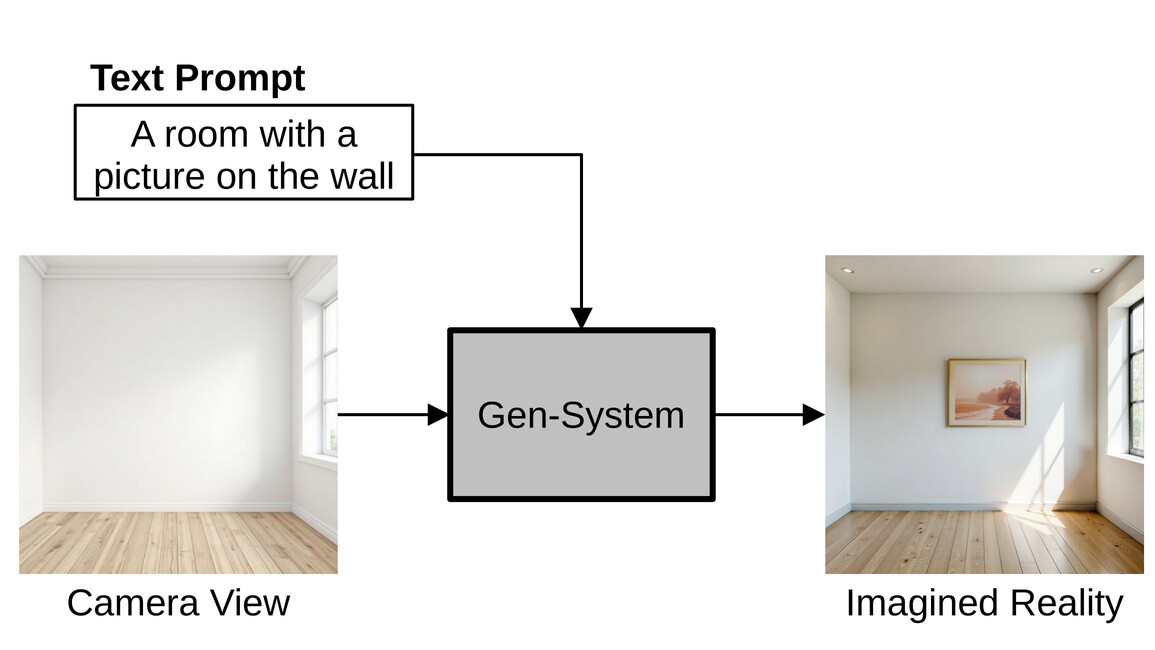}
  \caption{Overview of the overall system where a camera view of a robot is used in conjunction with a text-prompt to generate an imagined version of reality where the goal state is already achieved, here for hanging a painting on the wall (images generated with FLUX.1[dev] and FLUX.1 Canny[dev])}
  \label{fig:add_painting_to_wall}
\end{figure}

Assuming a robot was given the task of hanging a painting on the wall in an empty room and now is confronted with the problem of finding a suitable position while looking directly at the wall. We can use the generative system to generate an image based on the camera view where a painting already hangs on the wall, while preserving the overall layout of the scene. The idea of the system here is to create an imagined version of the reality in which the goal is already achieved. Figure \ref{fig:add_painting_to_wall} shows the basic idea of the system where the text prompt is used to add a new object to the scene that a potential robot is currently observing. 

We can use different image processing techniques, depending on the given tasks, after generating the goal image to extract the relevant information. We used YOLOv8~\cite{Yolo8Ultralytics}, specifically the YOLOv8x model that was trained on the Open Images V7 dataset~\cite{OpenImages}, to detect the positions of the added objects in the final image. Improved versions of YOLO exist~\cite{khanam2024yolov11}, but these models were mostly trained on the COCO dataset~\cite{lin2014microsoft}, lacking some of the object categories that we wanted to investigate. The image processing part can be additionally enhanced with more sensory information from the robot itself by using a depth camera and depth estimation to find the distance to objects in the scene or other sensors like RADAR or LIDAR when they are available on the robot.

The information generation part and the information extraction part can be arbitrarily swapped out depending on the task or resource constraints. In a resource constraint environment we might use a smaller and less capable model and different information extraction methods might be used depending on the potential tasks for the robot or the available sensors. However, the theoretical use case of these image generation models isn't limited to just object placement or arrangement as we could also use them to generate images of prototypical examples of objects, which can be used to guide further actions. In case of the previously mentioned example for creating a cheese sandwich, see \ref{fig:sandwich_example}, the robot could use an image generation model to create an example version, learning about what type of ingredients are used and how they are usually arranged. However, such use cases would require more complex image processing and understanding mechanisms to be able to incorporate such information into an actual decision making process. A robot could also generate an image of a cooking scene to determine which tools are needed for the task.

Our initial investigation was based on earlier StableDiffusion models, e.g. 1.5 or 2.1, in conjunction with the corresponding ControlNet~\cite{zhang2023adding} versions for the respective model. Improving image generation and prompt adherence capabilities of more modern systems, especially when it comes to generating the correct number of objects in a scene, led us to investigate FLUX.1[dev] and FLUX.1 Canny[dev]. 

\section{Experiments}

We implemented two experiments using the CoppeliaSim robot simulator~\cite{CoppeliaSim} to investigate potential use cases using a simulated robot arm. The goal was to find suitable positions for objects in two different environments. The robot arm acts based on the added information from the generative model and places the objects appropriately. We use the previously described method to add missing objects to the scene, as seen by camera inside the simulation, by conditioning the image generation process on the scene layout and adding the missing object using the text prompt. The first task is to place a bowl on a table, while the second task is to hang a painting on the wall inside a room.

\subsection{Experiment 1: Placing a Bowl on a Table}

\begin{figure}[ht]
  \centering
  \begin{subfigure}{.33\linewidth}
    \includegraphics[width=\linewidth]{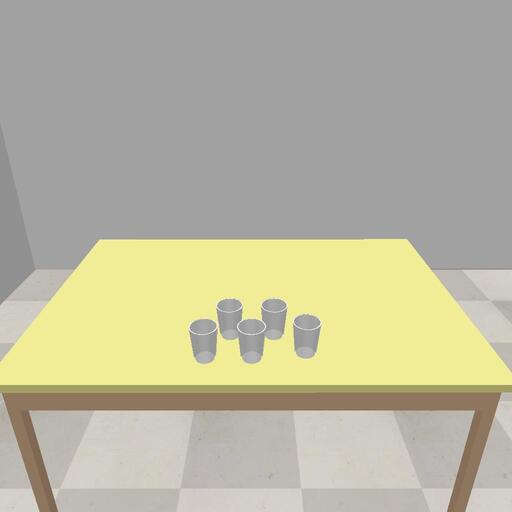}
    \caption{View in Simulation}
    \label{fig:bowl_placement_view}
  \end{subfigure}
  \begin{subfigure}{.33\linewidth}
    \includegraphics[width=\linewidth]{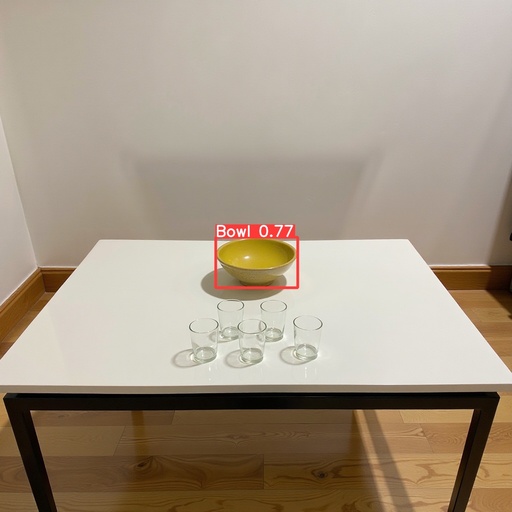}
    \caption{Generated Scene with Bounding Box}
    \label{fig:generated_bowl_image}
  \end{subfigure}
  \caption{Using the generative AI system to add a bowl to a table for a simulated scene. \textbf{(a)} Camera view inside the simulation, \textbf{(b)} Generated image with bowl and bounding box from YOLOv8 (generated with FLUX.1 Canny[dev])}
  \label{fig:bowl_scene_camera}
\end{figure}

The first experiment consists of a table in a room where parts of the table are occupied by glasses. The task is to place a bowl on a suitable spot on the table while the occupied parts should be avoided. The generative model is used to generate an image showing a bowl already on the table. This image is used to guide a robot arm inside the simulation to move the bowl from its starting position to its goal pose. Figure \ref{fig:bowl_scene_camera} shows the scene from the view of a camera placed inside the simulation and a generated image with the added bowl and the generated bounding box from YOLOv8.

We used FLUX.1[dev] and FLUX.1 Canny[dev] in conjunction with ComfyUI to generate the images and could already achieve good results using basic settings with a flux guidance score of 30, 20 inference steps with the euler sampler and the normal scheduler. However, the CFG score had to be increased from the default value of 1.0 to around 1.6 to reliably generate the additional objects in the image. Otherwise, FLUX would mostly ignore the added object in the prompt and create the image largely based on the edge map. A relatively simple prompt was used, broadly describing the scene and adding the missing object: "A room with a single bowl and glasses on a table".

\begin{figure}[ht]
	\centering
	\includegraphics[width=\linewidth]{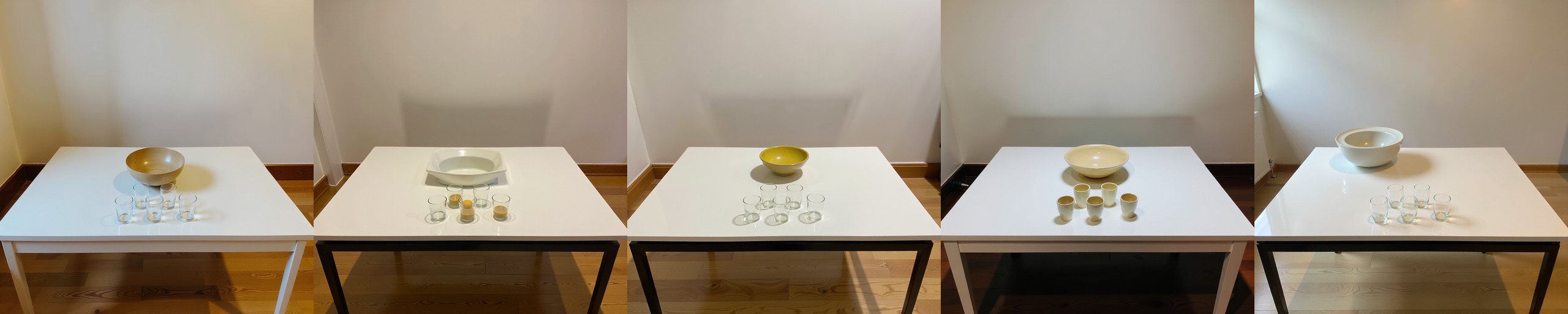}
	\caption{Multiple images generated for the bowl placing tasks from a batch generated using FLUX.1 Canny[dev], showing good consistency regarding the objects across all images.} 
    \label{fig:generated_bowl_scenes}
\end{figure}

Overall, the generative model showed good consistency when it comes to generating the correct objects from the prompt and edge map. Figure \ref{fig:generated_bowl_scenes} shows multiple images generated from a single batch. All images show a bowl placed at valid positions on the table and the correct number of glasses. 

\begin{figure}[ht]
  \centerline{\includegraphics[clip, width=\textwidth]{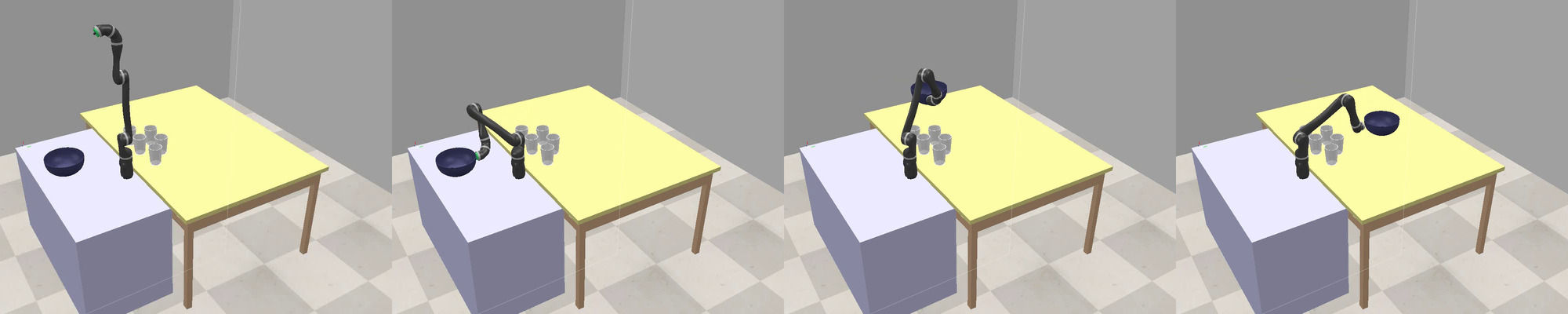}}
  \caption{Sequence of steps in the simulation for putting a bowl on the table: \textbf{a)} Starting Position \textbf{b)} Grasping the
Bowl \textbf{c)} Moving the bowl towards the goal \textbf{d)} Placing the bowl on the table}
  \label{fig:robot_bowl_movement}
\end{figure}

The position of the bowl on the table is approximated using a depth image from the simulated camera, giving us the distance from the camera for each pixel. The bottom center of the bounding box gives an estimate where the bowl hits the table and the depth value for this pixel allows us to estimate a distance from the camera to the imagined bowl. Additionally, using information from the camera, e.g. the perspective projection angle and image resolution, enables us to calculate a direction vector from the camera for each pixel. Combining direction vector and depth information gives us an approximate position of the imagined bowl on the table. However, a real robot would likely use a more refined method to estimate the 3D position of these objects. Figure \ref{fig:robot_bowl_movement} shows a simulated robot arm placing the bowl on the table based on a placement suggested by the generative system. The robot arm and bowl that can be seen are ignored by the camera for image generation.

\subsection{Experiment 2: Hanging a Picture on the Wall}

\begin{figure}[ht]
  \centering
  \begin{subfigure}{.33\linewidth}
    \includegraphics[width=\linewidth]{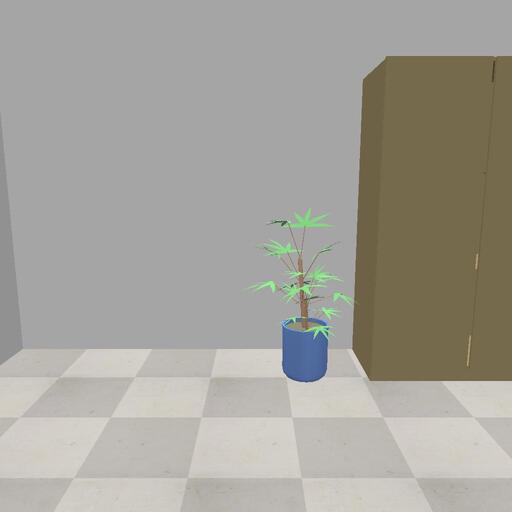}
    \caption{View in Simulation}
    \label{fig:frame_placement_view}
  \end{subfigure}
  \begin{subfigure}{.33\linewidth}
    \includegraphics[width=\linewidth]{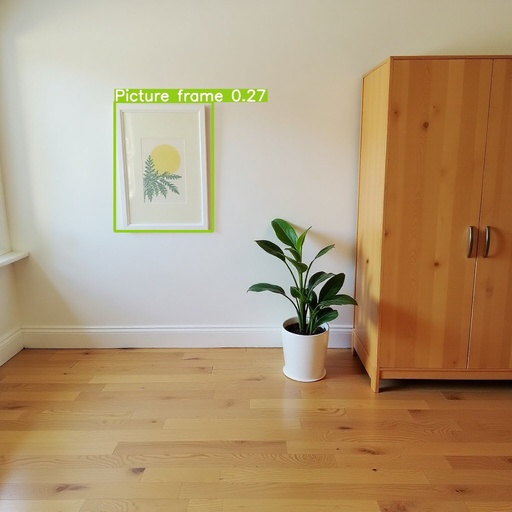}
    \caption{Generated Scene with Bounding Box}
    \label{fig:generated_frame_image}
  \end{subfigure}
  \caption{Using the generative AI system to add a picture on the wall for a simulated scene. \textbf{(a)} Camera view inside the
simulation, \textbf{(b)} Generated image with the picture frame and bounding box from YOLOv8 (generated with FLUX.1 Canny[dev])}
  \label{fig:frame_scene_camera}
\end{figure}

The second task is to put an image or picture frame on the wall. Here, the scene consists of wardrobe, a plant and an otherwise empty wall. The generative model is used to generate an image showing a picture inside a frame that already hangs on the wall. Figure \ref{fig:frame_scene_camera} shows the scene from the view of a camera placed inside the simulation and the generated image with the added picture frame on the wall. 

\begin{figure}[ht]
  \centerline{\includegraphics[clip, width=\textwidth]{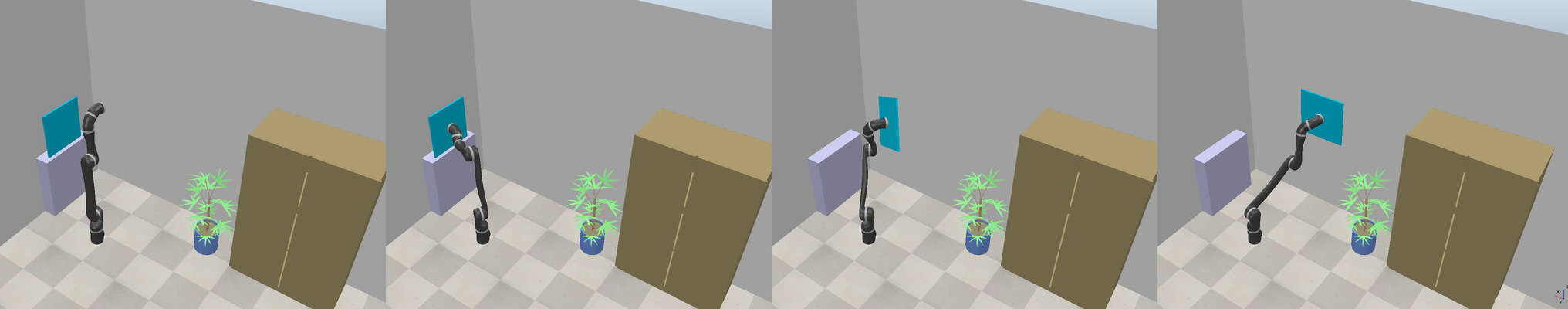}}
  \caption{Sequence of steps in the simulation for hanging a picture frame on the wall: \textbf{a)} Starting position \textbf{b)} Grasping the
picture frame \textbf{c)} Moving the picture frame towards the goal \textbf{d)} Placing the picture frame on the wall}
  \label{fig:robot_frame_movement}
\end{figure}

The position of the picture frame on the wall can be estimated based on the known height of the wardrobe and the length of the wall from the wardrobe to the left corner. Using the center of the bounding box allows us to estimate the center position of the image frame on the wall as shown in the image. Again, a real robot would likely use a more refined method to estimate the goal position. Figure \ref{fig:robot_frame_movement} shows a simulated robot arm placing a picture frame on the wall based on the proposed placement given by the generative model. The robot arm and picture frame, represented by the blue square, are ignored by the camera for the image generation part. Here we used the same base settings for the generative system with the prompts "A room with a plant, a cupboard and painting hanging on the wall" and "A room with a plant, a cupboard and a picture frame hanging on the wall" for generating corresponding images.

\section{Discussion}

In the future a single neural network might be used to receive the task in the form of a written assignment or as speech input when interacting with a human from a microphone, solve the task internally, and directly control the joints of the robot. In certain environments and for a restricted number of tasks, robots today are already capable of this. However, a full household robot that can handle a wide variety of tasks autonomously in different and changing environments is still not readily available. In regards to that, we showed that a pick-and-place task for a robot in a simulated environment can benefit from incorporating general world knowledge about object positioning from generative AI systems. However, the basic idea is more general and could be used for more complex scenarios under varying circumstances. Images of the world often contain information that is not necessarily written down or explicitly encoded but was incorporated or learned by humans while interacting with the world. In addition to object positions, it can give us prototypical examples of objects or meaningful object relations. Making this knowledge available to a robot by extracting knowledge from modern generative text-to-image systems can enable the robot to make better decisions. However, modern vision-language models are already capable of incorporating vision data into their reasoning mechanisms and might be able to handle such tasks directly without an explicit knowledge generation module. The future also might lie in directly combining textural and other sensory information directly into a single model, as can be seen by the development of modern Multi-Modal-LLMs\cite{zhang2024mm}, to directly guide the robot without introducing boundaries between the different domains.

\section{Conclusion}

We showed that modern image generation models can be used in different scenes and environments to guide robot action by extracting relevant information from generated images. These images contain knowledge about how humans would naturally place or arrange different objects. They can help envision prototypical examples of different objects. While earlier models had problems with generating the correct number of objects or were generating objects with unnatural artifacts, modern generative models are often able to generate correct scenes that can be very convincing to the viewer. Such systems could enhance the capabilities of robots in complex and previously unseen scenarios by drawing from the vast knowledge stored in images published on the web and making it easily usable to the current environment in which the robot has to act.

\section*{Acknowledgments}

The research reported in this paper has been supported by the German Research Foundation DFG, as part of the Collaborative Research Center (Sonderforschungsbereich) 1320 Project-ID 329551904 “EASE - Everyday Activity Science and Engineering”, University of Bremen (\url{http://www.ease-crc.org/}). The research was conducted as part of the subproject H01 “Sensorimotor and Causal Human Activity Models for Cognitive Architectures.”

\bibliographystyle{unsrt}  
\bibliography{robot_ai_world_knowledge}

\end{document}